\documentclass[journal]{IEEEtran}

\usepackage{amssymb} 
\usepackage[cmex10]{amsmath}
\usepackage{systeme}  
\usepackage{amsthm}  
\usepackage{stackengine}
\usepackage{titletoc} 
\usepackage{algorithmic}
\usepackage{algorithm}
\usepackage{array}
\usepackage{enumerate} 
\usepackage{enumitem}
\usepackage{verbatim}
\usepackage{graphicx}
\usepackage{graphics}
\usepackage{hyperref}
\usepackage[caption=false,font=normalsize,position=top]{subfig}
\usepackage[rightcaption]{sidecap} 
\graphicspath{ {./images/} }
\usepackage{url}
\usepackage{xcolor}
\usepackage{multirow}
\usepackage{tabularray}
\usepackage{arydshln} 
\usepackage{comment}
\usepackage{tikz,siunitx}
\usepackage{subfig}
\usepackage{booktabs, makecell, tabularx}
\usepackage{rotating}
\usepackage[export]{adjustbox}
\usepackage{authblk}
\usepackage{bm}

\begin{document}
    \title{Locality-preserving Directions for Interpreting  the Latent Space of  Satellite Image GANs }
    \author{Georgia Kourmouli*\thanks{*author contributed equally to this work}, Nikos Kostagiolas*, Yannis Panagakis, Mihalis A. Nicolaou}

    \maketitle

    \begin{abstract}
        We present a locality-aware method for interpreting the latent space of wavelet-based Generative Adversarial Networks (GANs),  that can well capture the large spatial and spectral variability that is characteristic to satellite imagery.  By focusing on  preserving locality, the proposed method is able to decompose the weight-space of pre-trained GANs and recover interpretable directions that correspond to high-level semantic concepts (such as urbanization, structure density, flora presence) - that can subsequently be used for guided synthesis of satellite imagery. In contrast to typically used approaches that focus on capturing the variability of the weight-space in a reduced dimensionality space (i.e., based on Principal Component Analysis, PCA), we show that  preserving locality leads to vectors with different angles, that are more robust to artifacts and can better preserve  class information.  Via a set of quantitative and qualitative examples, we further show that the proposed approach can outperform both baseline geometric augmentations, as well as global, PCA-based approaches for data synthesis in the context of data augmentation for satellite scene classification.
    \end{abstract}

    
    \IEEEpeerreviewmaketitle

    \section{Introduction}
    \IEEEPARstart{S}{ince} the advent of Generative Adversarial Networks (GANs) \cite{gans}, the field of synthetic image generation has been enjoying a series of unprecedented breakthroughs. Advances in the quality and fidelity of GAN generations, such as the introduction of progressive-growing \cite{progressivegans} and style-based architectures \cite{stylegan,stylegan2}, have recently become the norm. Despite these models largely focusing on spatial information for their learning, the benefits of augmenting image-specific models with spectral information have also been argued for in the past \cite{wavelet1,wavelet3,wavelet8}, mainly due to their ability to potentially tackle the spectral bias inherent in GANs. Indeed, end-to-end frequency-based generative models have recently yielded promising results, not only in the natural image space \cite{swagan} but also in the more demanding, high-resolution, birds-eye-view setting of satellite imagery data \cite{nikou}. 
    
    \looseness-1Owing to the hierarchical nature of their representations, GAN-like architectures, and their variants are able to capture a wide range of semantics both in their intermediate feature \cite{gandissection} and latent space \cite{ganalyze,steerability,interpreting}. In this light, both supervised and unsupervised efforts aiming towards inferring these semantics soon emerged \cite{ganspace,unsupervised1,sefa,hessian,stylespace}, thus paving the way towards ad-hoc latent semantic exploration. 
    
    A popular way of addressing the problem of semantic exploration in GANs is applying a dimensionality reduction algorithm (e.g., Principal Component Analysis, PCA) on the weight matrix of their projection layers, in order to expose directions that cause large variations in the output images. This strategy has been successfully applied to a large variety of real-world image categories \cite{sefa}, including satellite images \cite{nikou}. Edits across a large variety of distinct directions corresponding to high-level semantic concepts are enabled, particularly in domains such as \textit{gender} and \textit{age} in the former or \textit{urbanization} and \textit{vegetation growth} in the latter. 
    An alternative to global methods, such as PCA, that focus on finding the directions of maximum variance, is to regularize the dimensionality reduction process by seeking also to preserve the local data structure \cite{lpp}. When introduced to the interpretation of generative networks,  locality-based constraints can improve the generation process by facilitating  more localized edits, a property that is beyond the scope of earlier trials based on maximizing variance. Applications like data augmentation which rely heavily on data variability, have much to gain from such improvements, rendering them a natural candidate for any trials along this axis.
    
    In light of those recent advances in generative modeling as well as prior ones in the field of dimensionality reduction, we present in this study the first case, to our knowledge, of incorporating the Locality Preserving Projections \cite{lpp} algorithm into the latent space exploration pipeline of a wavelet-based GAN \cite{swagan}. We hypothesize that locality preservation plays a pivotal role in strengthening the diversity of a generative model's generations, especially when it comes to satellite image data. Satellite image datasets that naturally consist of higher spatial and spectral variability constitute a more challenging medium than natural images. 
    Our work extends earlier trials focusing on bridging PCA-based latent exploration techniques with satellite image applications \cite{nikou}. Through an extensive series of data augmentation experiments and related ablation studies, we show that locality-aware methods can outperform their PCA-based counterparts in downstream tasks. Our study's main contributions can be summarized as follows:
    \begin{itemize}[noitemsep,topsep=0pt]
        \item We propose the first locality-aware approach for interpreting the latent space of GANs trained on satellite images by discovering meaningful semantic directions.
        \item With a series of qualitative experiments, we demonstrate the method's efficacy in generating diverse images that respect the underlying local structure of the data.
        \item  Moreover, we present a rigorous set of quantitative experiments that demonstrate the favourable properties of preserving the local  structure.  Notably, the proposed method outperforms compared data augmentation approaches (optimization-based or geometric) for remote scene classification, even on {\it unseen datasets}. 
        \item  Finally, we demonstrate that the proposed approach can readily be used in a complimentary fashion with geometric-based augmentations, further improving accuracy for downstream tasks.
    \end{itemize}

    \section{Methodology}
    \subsection{Preliminaries}
    Earlier work on latent semantic exploration \cite{sefa, ganspace}, also applied on the satellite images \cite{nikou},  discovers  directions that correspond to high-level semantic concepts that are present in satellite imagery. This is achieved by analyzing the weight-space of a generative model, namely the weight matrix $\mathbf{A}$ corresponding to projection layers of the aforementioned generative model.  In particular, in \cite{sefa}, a Principal Component (PCA) based approach is presented, that recovers the directions that lead to large output variation - corresponding to the top-$k$ eigenvectors $\mathbf{u}_i$ (corresponding in turn to the $k$ eigenvalues $n_i$)  of the covariance matrix of weights, $\mathbf{A}^T\mathbf{A}$.

    \begin{figure*}
    \centering
        \includegraphics[width=\linewidth]{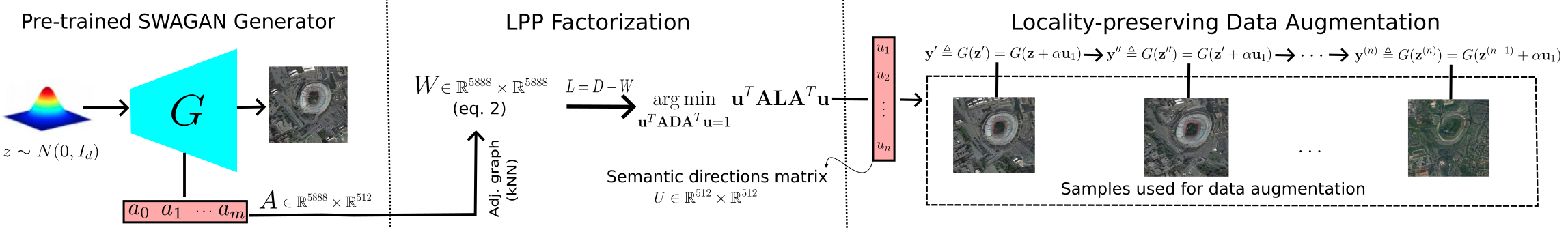}
        \caption{Overview of our method, including the various steps (both in terms of dimensionality and processing) our data goes through.}
        \label{fig:overview}
    \end{figure*}
    
    Given these directions, edits on the original image can be achieved by manipulating its latent code $\mathbf{z}$ towards the direction $\mathbf{u}_i$ that corresponds to the target concept by $\mathbf{z}' = \mathbf{z}+\alpha\mathbf{u}_i$, where $\alpha$ is a scalar that stands for the magnitude of the edit. The edited image $\mathbf{y}'$ can be obtained by
    \begin{equation} \label{eq:edit}
    \mathbf{y}' \triangleq G(\mathbf{z}')=G( \mathbf{z}+\mathbin{\alpha}\mathbf{u}_i) ,  
    \end{equation}
    where $G$ denotes the application of the generator model.

    \subsection{Discovering Locality-preserving Interpretable Directions} \label{sub:method}
    We aim to discover directions that preserve the local structure of the GAN latent space, while still relying on the hypothesis that the weight space of the generator lies on a low-dimensional manifold.  In the linear sense, the optimal approach for achieving this is Locality Preserving Projections (LPP) \cite{lpp}.  Following LPP, instead of working on the covariance of the weight matrix, we propose to adopt an approach that works on the adjacency graph of the generator weights.  In particular, we find an adjacency graph by applying $k$-nearest neighbours on the weight vectors of  the generator, $\mathbf{A}=[\mathbf{a}_{0},\ldots,\mathbf{a}_{m}]$. We then construct the weight matrix $\mathbf{W}$ as:
    \begin{equation}
    	\mathbf{W}_{ij} = 
    	\begin{cases}
    		1, & \text{if $\mathbf{a}_{i}$, $\mathbf{a}_{j}$ neighbours}\\
    		0, & \text{otherwise}
    	\end{cases} 
    \end{equation} 
    The optimal linear mapping of this graph, so that connected points stay as close as possible, can be found by solving the following generalized eigenvalue problem  \cite{lpp}
    \begin{equation} \label{eq:opt-problem}
        \underset{\mathbf{u}^{T}\mathbf{A}\mathbf{DA}^{T}\mathbf{u} = 1}{\arg\min} \mathbf{u}^{T}\mathbf{A}\mathbf{LA}^{T}\mathbf{u},
    \end{equation}
    where $\mathbf{A}$ is the model's weight matrix, $\mathbf{D}$ is a diagonal degree matrix of $\mathbf{W}$ and $\mathbf{L} = \mathbf{D}-\mathbf{W}$ the Laplacian matrix.

    Having obtained  directions that correspond to the top $k$-eigenvectors of  Eq. \ref{eq:opt-problem}, we can readily apply Eq. \ref{eq:edit} to edit based on the recovered interpretable dimensions that preserve local structure, and can  better preserve neighbourhood information.  Furthermore, the proposed approach is a generalization of PCA-based approaches, as constructing a neighbourhood graph where all points are connected reduces to PCA.

    \subsection{Experimental Setting}
    \looseness-1To discover the semantic directions with the proposed method, we utilise a SWAGAN  checkpoint\footnote{https://github.com/kostagiolasn/SatConcepts}, pre-trained on Resisc-45. We subsequently use the weight matrix $\mathbf{A} \in \mathbb{R}^{5888}\times\mathbb{R}^{512}$ of the SWAGAN model and the $k=10$ nearest neighbours of each weight, to build the adjacency graph and subsequently the matrices $\mathbf{W,D}$ and $\mathbf{L}$. Solving the generalized eigenvalue problem in Eq. \ref{eq:opt-problem}, we keep the first 512 components. We find that the top-7 eigenvectors represent the most significant directions in the latent space, similarly to \cite{nikou}. 

    In order to perform an edit on the images produced by SWAGAN with the directions found by our method, we use the resulting eigenvectors $\mathbf{u}_i$ with the original latent codes $\mathbf{z} \in \mathbb{R}^{512}$ as presented in Eq. \ref{eq:edit}. This process, detailed in Fig. \ref{fig:overview}, follows the pattern of the respective PCA implementation in GANs \cite{sefa} and particularly the one that was used in \cite{nikou}. 

    As shown in \cite{nikou}, the implementation of PCA in GANs can also be used in downstream tasks like data augmentation. In order to further evaluate our method, we performed a series of data augmentation experiments and compared their results with those produced by other techniques. In particular, we use three different datasets containing satellite images, that we artificially imbalance by reducing their training sizes and three techniques to perform the data augmentation:
    \begin{enumerate}[noitemsep,topsep=0pt]
        \item \textbf{Geometric transformations:} We apply three random rotations sampled from the set $\{30^{\circ}, 60^{\circ}, 90^{\circ}, 120^{\circ},$ $150^{\circ}, 210^{\circ}, 240^{\circ}, 270^{\circ}\}$ and one horizontal flip, producing 4 extra images for every sample in the training set. This method is denoted as Baseline in our experiments.
        \item \textbf{PCA-based augmentation:} We generate a synthetic sample with SWAGAN and by applying to it the first eigenvector found by PCA we produce 4 augmented ones using $d=\pm 1$ degree of freedom in movement and $\pm 1,$ $\pm 2$ steps along the corresponding direction (unless stated otherwise). We then classify the augmented results using a ResNet-50 classifier \cite{resnet50} that was trained on Resisc-45.
        \item \textbf{LPP-based augmentation:} We follow the process described in the PCA-based augmentation, which suggests the usage of the first eigenvector found by LPP and the magnitudes $\alpha=\pm 1, \pm 2$ to produce 4 extra results. The augmented results are then classified by ResNet-50.
    \end{enumerate}
    
    We compare these approaches on the task of data augmentation for remote scene classification in multiple settings, both on the whole test set (denoted as Acc. in experiments) and on a test set with only the classes that were artificially imbalanced (denoted as Imb. Acc. in experiments). 

    \subsection{Datasets}
    The image scene classification and augmentation experiments used variations of three publicly available datasets containing satellite images. The NWPU-RESISC45 \cite{resisc45} which consists 31.500 images from 45 scene classes, the AID \cite{aid} with 10.000 images from 30 scene classes and the UCMerced \cite{ucmerced} with 2100 images from 21 scene classes. To evaluate and compare the performance of the various data augmentation techniques, we created five datasets by artificially unbalancing the aforementioned ones, as shown in Table \ref{table:dataset-variations}.
    
    \begin{table}
        \caption{Dataset Variations used in Experiments}
        \label{table:dataset-variations}
        \centering
        \begin{tblr}{
          rowspec={Q[m]Q[m]Q[m]Q[m]Q[m]},
          colspec = {Q[218]Q[123]Q[123]Q[123]Q[194]Q[86]},
          rowsep = 1pt,
          cell{2}{2} = {c=3}{0.276\linewidth,c},
          cell{2}{5} = {c},
          cell{2}{6} = {c},
          cell{3}{2} = {c},
          cell{3}{3} = {c},
          cell{3}{4} = {c},
          cell{3}{5} = {c},
          cell{3}{6} = {c},
          cell{4}{2} = {c=3}{0.276\linewidth,c},
          cell{4}{5} = {c},
          cell{4}{6} = {c},
        }
        \SetHline{1-6}{1.5pt}
        Dataset   & Resisc70 & Resisc35 & Resisc10 & UCMerced10 & AID40 \\ 
        \hline
        No. of imb. classes   & 7        &          &          & 5          &  7 \\
        Train size per imb./balanced class  & 70/450  & 35/450  & 10/450  & 10/75  & 40/120\\
        Validation/test ratio per class  & 150/100  &   &   & 15/10  & 40/40 \\
        \SetHline{1-6}{1.5pt}
       \end{tblr}
    \end{table}    
    
    \section{Experiments}    
    \subsection{Controllable Image Synthesis via Intepretable Directions}
    In this set of experiments, we used the proposed method described (Sec. \ref{sub:method}), to search for semantically meaningful directions in the latent space of SWAGAN. The directions found (Table \ref{table:directions}) semantically correspond to similar changes in the image as PCA \cite{nikou}.  However, the directions are rotated based on the connectivity of the weight vector graph, thus enabling editing in a locality-aware fashion. Their deviation (Table \ref{table:directions}-\textbf{Angle}) demonstrates that the directions found by the two methods are indeed different, in agreement to \cite{lpp}.
     
\begin{figure*}
      \centering
        \tikz[baseline]{\draw (0,0.55) node [rotate=90] [inner sep=0] {\scriptsize{Locality-}};}
        \tikz[baseline]{\draw (0,0.5) node [rotate=90] [inner sep=0] {\scriptsize{based}};}
        \includegraphics[angle=90,width=0.7\linewidth]{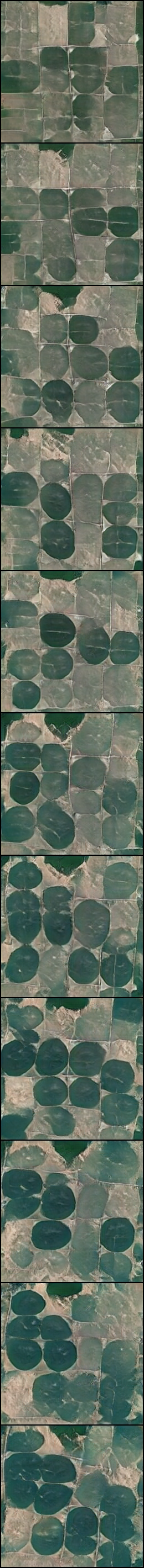}\\
        \tikz[baseline]{\draw (0,0.6) node [rotate=90] [inner sep=0] {\scriptsize{PCA-based}};}
        \tikz[baseline]{\draw (0,0.1) node [rotate=90] [inner sep=0.3mm] {\scriptsize{ }};}
        \tikz[baseline]{\draw (0,0.1) node [rotate=90] [inner sep=0] {\scriptsize{ }};}
        \includegraphics[angle=90,width=0.7\linewidth]{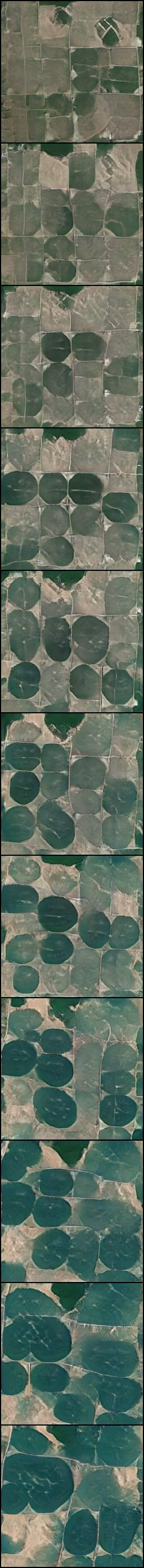}

        \tikz[baseline]{\draw[line width=0.4mm] (0,0.4)--(13.8,0.4) ;}
        \vspace*{-2.4ex}

        \tikz[baseline]{\draw (0,0.55) node [rotate=90] [inner sep=0] {\scriptsize{Locality-}};}
        \tikz[baseline]{\draw (0,0.5) node [rotate=90] [inner sep=0] {\scriptsize{based}};}
        \includegraphics[angle=90,width=0.7\linewidth]{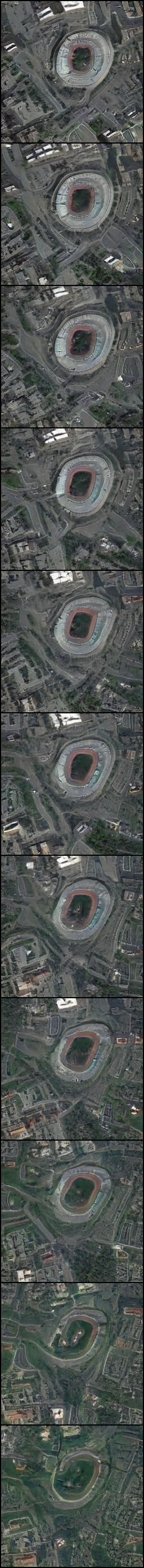} \\ 
        \tikz[baseline]{\draw (0,0.58) node [rotate=90] [inner sep=0] {\scriptsize{PCA-based}};}
        \tikz[baseline]{\draw (0,0.1) node [rotate=90] [inner sep=0.3mm] {\scriptsize{ }};}
        \tikz[baseline]{\draw (0,0.1) node [rotate=90] [inner sep=0] {\scriptsize{ }};}
        \includegraphics[angle=90,width=0.7\linewidth]{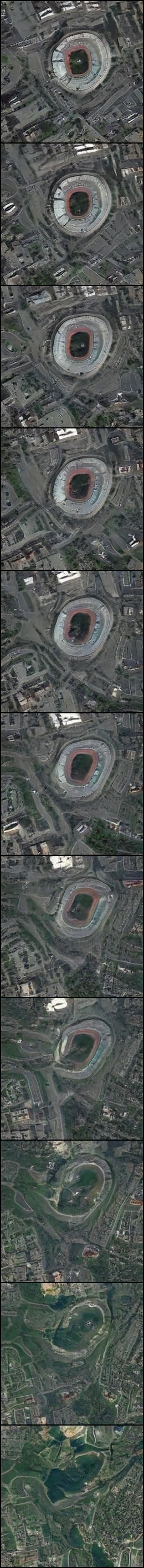}
        
        \tikz[baseline]{\draw[line width=0.4mm] (0,0.4)--(13.8,0.4) ;}
        \vspace*{-2.4ex}

        \tikz[baseline]{\draw (0,0.55) node [rotate=90] [inner sep=0] {\scriptsize{Locality-}};}
        \tikz[baseline]{\draw (0,0.5) node [rotate=90] [inner sep=0] {\scriptsize{based}};}
        \includegraphics[angle=90,width=0.7\linewidth]{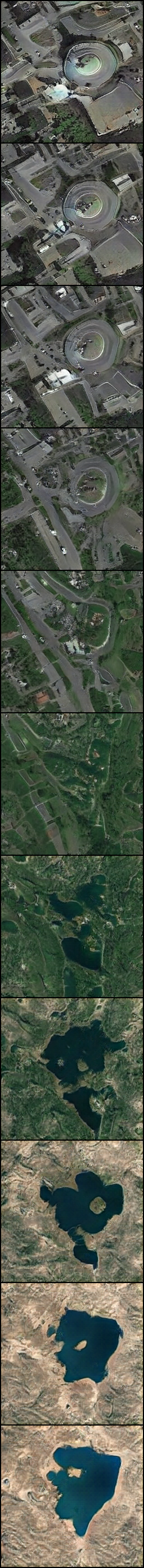}\\
        \tikz[baseline]{\draw (0,0.58) node [rotate=90] [inner sep=0] {\scriptsize{PCA-based}};}
        \tikz[baseline]{\draw (0,0) node [rotate=90] [inner sep=0.3mm] {\scriptsize{ }};}
        \tikz[baseline]{\draw (0,0) node [rotate=90] [inner sep=0] {\scriptsize{ }};}
        \includegraphics[angle=90,width=0.7\linewidth]{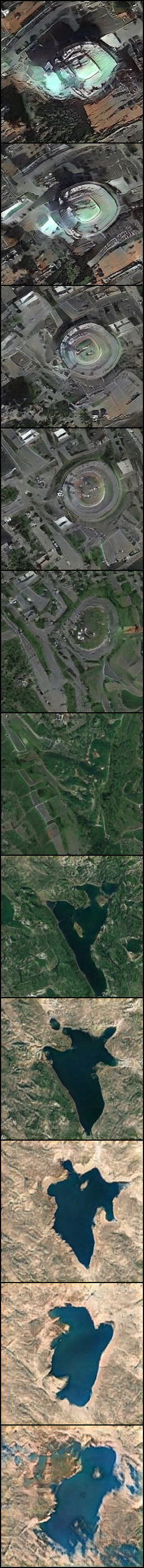}\\[-0.2ex]

        \tikz[baseline]{\draw[line width=0.4mm] (0,0.4)--(13.8,0.4) ;}
        \vspace*{-2.4ex}

        \tikz[baseline]{\draw (0,0.55) node [rotate=90] [inner sep=0] {\scriptsize{Locality-}};}
        \tikz[baseline]{\draw (0,0.5) node [rotate=90] [inner sep=0] {\scriptsize{based}};}
        \includegraphics[angle=90,width=0.7\linewidth]{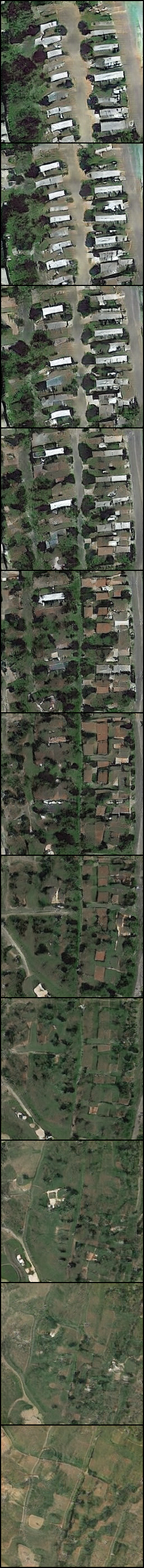}\\[-0.2ex]
        \tikz[baseline]{\draw (0,0.58) node [rotate=90] [inner sep=0] {\scriptsize{PCA-based}};}
        \tikz[baseline]{\draw (0,0) node [rotate=90] [inner sep=0.3mm] {\scriptsize{ }};}
        \tikz[baseline]{\draw (0,0) node [rotate=90] [inner sep=0] {\scriptsize{ }};}
        \includegraphics[angle=90,width=0.7\linewidth]{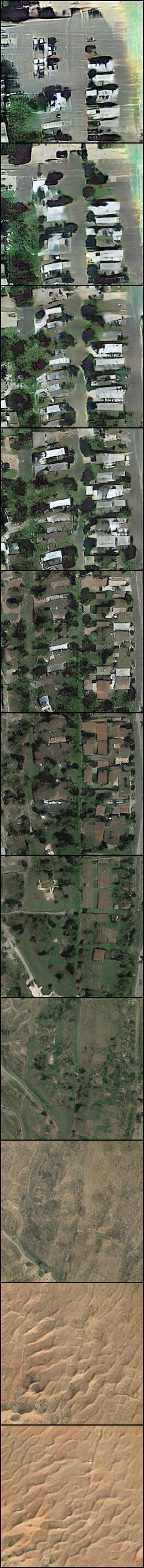}
        
        \vspace*{-0.3ex}
        \caption{Comparison between Locality- and PCA-based synthesis with magnitudes $1.0, 1.0, 3.0$ and $3.0$ for every respective set of images.}
        \label{fig:comparison4}
    \end{figure*}
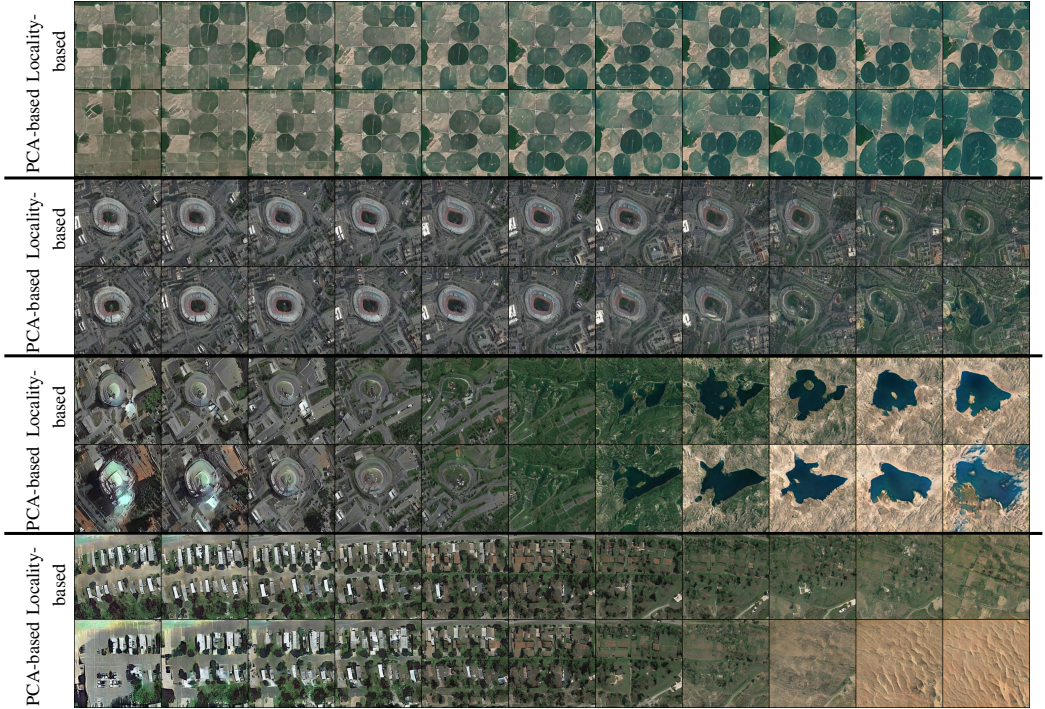

    In general, the first direction corresponding to the largest eigenvalue introduces urbanization concepts strongly, even in images representing areas of natural scenery. The less important directions stemming from smaller eigenvalues, tend to alter some features upon the image, as indicated in results and Table \ref{table:directions}. 
    Observing the results in Fig. \ref{fig:comparison4}, it is evident that the edits performed using the locality-based method are better both in quality and class-preservation. In the 1st and 2nd set of images, the locality-based method manages to maintain the shapes and classes created in the images, while the PCA-based one fails to do so despite the small magnitude used. In the last two set of images, it is visible that the PCA-based approach creates artifacts in all extreme cases while the Locality-based one maintains the overall structure.
    
    \begin{table}
        \caption{Semantic Directions with their corresponding edit description}
        \label{table:directions}
        \centering
        \begin{tblr}{
          rowspec={Q[m]Q[m]Q[m]Q[m]Q[m]Q[m]Q[m]},
          colspec = {Q[100]Q[180]Q[200]Q[40]},
          rowsep = 1pt,
        }
        \SetHline{1-4}{1.5pt}
        \textbf{Semantic Direction} \# & \textbf{Description (+)}  & \textbf{Description (-)} & \textbf{Angle}\\
        \hline
        1  & Urbanization Growth & Urbanization Decrease & 47.32$^{\circ}$\\ 
        3  & Vertical axis structure reinforcement & Horizontal axis structure reinforcement & 58.35$^{\circ}$\\ 
        4  & Secondary structure growth & Main structure growth & 60.73$^{\circ}$\\
        5  & Road structure intro/ reinforcement & Road structure removal & 48.05$^{\circ}$\\
        7  & Flora decrease upon existing structure & Flora growth upon existing structure & 41.95$^{\circ}$\\
        \SetHline{1-4}{1.5pt}
        \end{tblr}
        \par
        \vspace{1ex}
        {\raggedright \textit{\textbf{Note:} We also specify the angle between the corresponding eigenvectors found by LPP- and PCA-based interpretations.} \par}
    \end{table}
    
    \subsection{Data Augmentation via Intepretable Directions}
    \subsubsection{Experiment I}
\begin{figure}
        \centering
        \tikz[baseline]{\draw (0,-0.5) node [rotate=90] [inner sep=0] {\scriptsize{Locality-}};}
        \tikz[baseline]{\draw (0,-0.5) node [rotate=90] [inner sep=0] {\scriptsize{based}};}
        \subfloat[]{\includegraphics[width=0.12\linewidth]{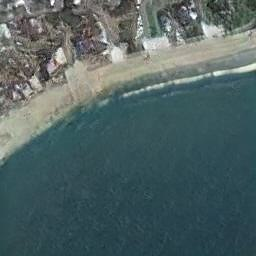}%
        }
        \hspace{0.01mm}
        \subfloat[]{\includegraphics[width=0.12\linewidth]{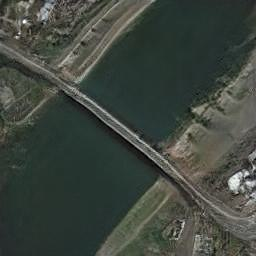}%
        }
        \hspace{0.01mm}
        \subfloat[]{\includegraphics[width=0.12\linewidth]{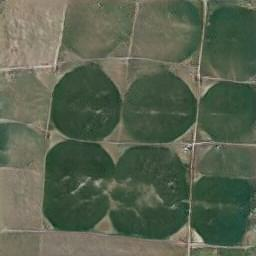}%
        }
        \hspace{0.01mm}
        \subfloat[]{\includegraphics[width=0.12\linewidth]{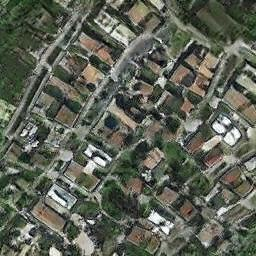}%
        }
        \hspace{0.01mm}
        \subfloat[]{\includegraphics[width=0.12\linewidth]{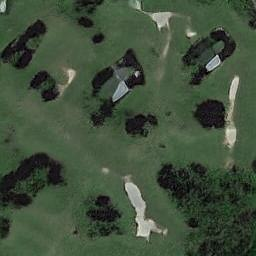}%
        }
        \hspace{0.01mm}
        \subfloat[]{\includegraphics[width=0.12\linewidth]{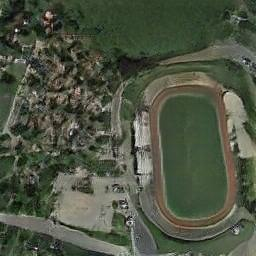}%
        }
        \hspace{0.01mm}
        \subfloat[]{\includegraphics[width=0.12\linewidth]{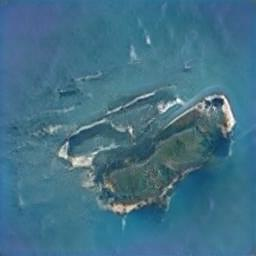}%
        }\\[-1.7ex]
        \tikz[baseline]{\draw (0,-0.45) node [rotate=90] [inner sep=0.3mm] {\scriptsize{PCA-}};}
        \tikz[baseline]{\draw (0,-0.5) node [rotate=90] [inner sep=0] {\scriptsize{ based}};}
        \subfloat{\includegraphics[width=0.12\linewidth]{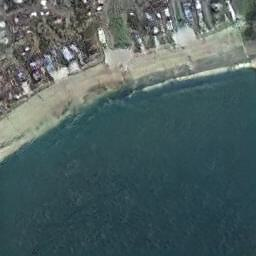}%
        }
        \hspace{0.01mm}
        \subfloat{\includegraphics[width=0.12\linewidth]{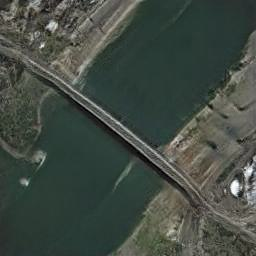}%
        \label{bridge}}
        \hspace{0.01mm}
        \subfloat{\includegraphics[width=0.12\linewidth]{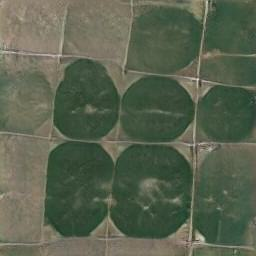}%
        \label{circular_farmland}}
        \hspace{0.01mm}
        \subfloat{\includegraphics[width=0.12\linewidth]{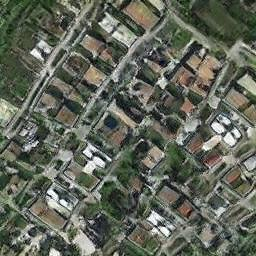}%
        \label{dense_residential}}
        \hspace{0.01mm}
        \subfloat{\includegraphics[width=0.12\linewidth]{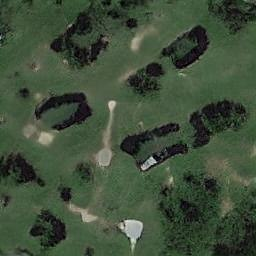}%
        \label{golf_course}}
        \hspace{0.01mm}
        \subfloat{\includegraphics[width=0.12\linewidth]{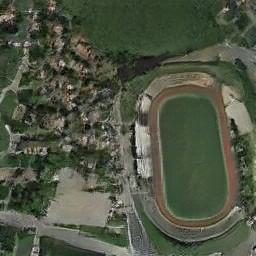}%
        \label{ground_track_field}}
        \hspace{0.01mm}
        \subfloat{\includegraphics[width=0.12\linewidth]{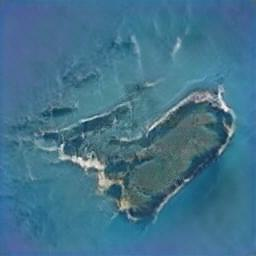}%
        \label{island}}
        \caption{Example of edits (using the first eigenvector) that were used to augment the Resisc-45 imbalanced classes in Experiment I: (a) \textit{beach}, (b) \textit{bridge}, (c) \textit{circular\_farmland}, (d) \textit{dense\_residential}, (e) \textit{golf\_course}, (f) \textit{ground\_track\_field}, (g) \textit{island}.}
        \label{fig:data_augmentation_edits}
    \end{figure}

    \looseness-1This experiment aims at comparing the different data augmentation techniques. The augmented results were classified using ResNet-50 with a filter of 0.8, on the highest class probability. We opted to increase the training size of the datasets using the methods Baseline, PCA-based and LPP-based by 5, and using the mixed methods by 9. Example of images that were used in this experiment can be found in Fig. \ref{fig:data_augmentation_edits}.
    
    Results of  data augmentation for remote scene classification can be found in Tables \ref{table:experiment1-results} and \ref{table:experiment1-mixedresults}a. By observing the results, it appears that the LPP-based method creates more diverse images, which contribute to a higher variability inside the imbalanced classes of the datasets. This is illustrated by the higher accuracies produced by this method across all the settings indicated in the table. Interestingly, as can be seen in Table \ref{table:experiment1-mixedresults}a, the combination of geometric and LPP-based augmentations outperforms the individual methods in Table \ref{table:experiment1-results}. Thus, it is demonstrated that the proposed method can be used in a complementary fashion to traditional augmentation approaches for improved results.
    
    \begin{table*}[t]
        \caption{Experiment I Results}
        \label{table:experiment1-results}
        \centering
        \begin{tabular}{p{1.4cm}cccccccc} 
        \specialrule{1.5pt}{1pt}{1.5pt} 
        \textbf{Dataset}
         & \multicolumn{2}{c}{\textbf{Without Augmentation}}      
         & \multicolumn{2}{c}{\textbf{Baseline}} 
         & \multicolumn{2}{c}{\textbf{PCA-based}}       
         & \multicolumn{2}{c}{\textbf{LPP-based}} \\
        \cmidrule(lr){2-3}\cmidrule(lr){4-5}\cmidrule(lr){6-7}\cmidrule(lr){8-9}
        \textbf{Variant} 
        & \textbf{Acc.}  & \textbf{Imb. Acc.}
        & \textbf{Acc.}  & \textbf{Imb. Acc.} 
        & \textbf{Acc.}  & \textbf{Imb. Acc.} 
        & \textbf{Acc.}  & \textbf{Imb. Acc.}\\ 
        \midrule
        Resisc-70       
        & 0.888 $\pm$ 0.005   & 0.789 $\pm$ 0.018  
        & 0.895 $\pm$ 0.004 & 0.843 $\pm$ 0.028   
        & 0.896 $\pm$ 0.016 & 0.844 $\pm$ 0.014   
        & \textbf{0.904 $\bm{\pm}$ 0.009} & \textbf{0.856 $\bm{\pm}$ 0.008}\\ 
        Resisc-35       
        & 0.868 $\pm$ 0.007   & 0.681 $\pm$ 0.019  
        & 0.894 $\pm$ 0.010 & 0.782 $\pm$ 0.012  
        & 0.887 $\pm$ 0.003 & 0.768 $\pm$ 0.021   
        & \textbf{0.899 $\bm{\pm}$ 0.012} & \textbf{0.796 $\bm{\pm}$ 0.011} \\ 
        Resisc-10      
        & 0.827 $\pm$ 0.006   & 0.412 $\pm$ 0.033  
        & 0.847 $\pm$ 0.007 & 0.543 $\pm$ 0.011   
        & 0.855 $\pm$ 0.011 & 0.587 $\pm$ 0.045   
        & \textbf{0.868 $\bm{\pm}$ 0.007} & \textbf{0.611 $\bm{\pm}$ 0.025}\\ 
        UCMerced10     
        & 0.878 $\pm$ 0.013 & 0.636 $\pm$ 0.053   
        & 0.909 $\pm$ 0.011 & 0.772 $\pm$ 0.037   
        & 0.888 $\pm$ 0.019 & 0.732 $\pm$ 0.057   
        & \textbf{0.915 $\bm{\pm}$ 0.014} & \textbf{0.784 $\bm{\pm}$ 0.032}\\
        AID-40          
        & 0.877 $\pm$ 0.013 & 0.800 $\pm$ 0.032  
        & 0.890 $\pm$ 0.005 & 0.847 $\pm$ 0.007  
        & 0.890 $\pm$ 0.020 & 0.853 $\pm$ 0.017  
        & \textbf{0.891 $\bm{\pm}$ 0.010} & \textbf{0.862 $\bm{\pm}$ 0.027}\\
        \specialrule{1.5pt}{1pt}{1pt}
        \end{tabular}
        \par
        \vspace{1ex}
        {\raggedright \textit{\textbf{Note:} The mean accuracy on the whole test set is denoted by Acc. and the one on the test set with only the imbalanced classes by Imb. Acc.} \par}
    \end{table*}
	
  	\begin{table*}[t]
  		\caption{Follow-up Experiments}
  		\centering
  		\label{table:experiment1-mixedresults}
  		\begin{tabular}{p{1.4cm}cccc}
  			\multicolumn{5}{c}{\small{(a) Experiment I Mixed Methods Results}}\\[1.2mm] 
  			\specialrule{1.5pt}{1pt}{1.5pt}
  			\textbf{Dataset}
  			& \multicolumn{2}{c}{\textbf{Mixed PCA-based}} 
  			& \multicolumn{2}{c}{\textbf{Mixed LPP-based}} \\
  			\cmidrule(lr){2-3}\cmidrule(lr){4-5}
  			\textbf{Variant} 
  			& \textbf{Acc.}  & \textbf{Imb. Acc.} 
  			& \textbf{Acc.}  & \textbf{Imb. Acc.}  \\ 
  			\midrule
  			Resisc-70    
  			& .911 $\pm$ .005 & .894 $\pm$ .005    
  			& \textbf{.914 $\bm{\pm}$ .006} & \textbf{.898 $\bm{\pm}$ .007} \\ 
  			Resisc-35    
  			& \textbf{.894 $\bm{\pm}$ .014} & .832 $\pm$ .019    
  			& .891 $\pm$  .009 & \textbf{.834 $\bm{\pm}$ .032} \\ 
  			Resisc-10    
  			& .866 $\pm$ .015 & .648 $\pm$ .028    
  			& \textbf{.874 $\bm{\pm}$ .014} & \textbf{.683 $\bm{\pm}$ .041} \\ 
  			UCMerced10  
  			& .914 $\pm$ .019 & .784 $\pm$ .050    
  			& \textbf{.923 $\bm{\pm}$ .024} & \textbf{.792 $\bm{\pm}$ .057} \\
  			AID-40       
  			& .898 $\pm$ .005 & .861 $\pm$ .030    
  			& \textbf{.898 $\bm{\pm}$ .009} & \textbf{.877 $\bm{\pm}$ .021}\\ 
  			\specialrule{1.5pt}{1pt}{1pt}
  		\end{tabular} 
  		\qquad
  		\label{table:experiment3+4-baseline}
  		\begin{tabular}{cc}
  			\multicolumn{2}{c}{\small{(b) Experiment III \& IV}}\\[1.2mm] 
  			\specialrule{1.5pt}{1pt}{1.5pt}
  			\multicolumn{2}{c}{\textbf{Baseline}}\\
  			\midrule
  			\textbf{Acc.}  & \textbf{Imb. Acc.}\\
  			\midrule
  			.909 $\pm$ .003 & .894 $\pm$ .004\\ 
  			.898 $\pm$ .006 & .842 $\pm$ .007\\
  			.856 $\pm$ .013 & .574 $\pm$ .031\\
  			.920 $\pm$ .022 & .788 $\pm$ .045\\
  			.884 $\pm$ .018 & .854 $\pm$ .045\\ 
  			\specialrule{1.5pt}{1pt}{1pt}
  		\end{tabular} 
  		\qquad
  		\label{table:experiment5-results}
  		\begin{tabular}{cc}
  			\multicolumn{2}{c}{\small{(c) Experiment V Results}}\\[1.2mm]
  			\specialrule{1.5pt}{1pt}{1.5pt}
  			\textbf{PCA-based}      & \textbf{LPP-based}\\
  			\midrule
  			\textbf{Imb. Acc.} & \textbf{Imb. Acc.} \\
  			\midrule
  			.819 $\pm$ .018   & \textbf{.838 $\bm{\pm}$ .015} \\ 
  			.749 $\pm$ .025   & \textbf{.767 $\bm{\pm}$ .034}\\
  			.546 $\pm$ .024   & \textbf{.578 $\bm{\pm}$ .039} \\
  			.640 $\pm$ .052   & \textbf{.688 $\bm{\pm}$ .055} \\
  			\textbf{.861 $\bm{\pm}$ .029}   & .839 $\pm$ .037 \\ 
  			\specialrule{1.5pt}{1pt}{1pt}
  		\end{tabular}
  		\par
  		\vspace{1ex}
  		{\raggedright \textit{\textbf{Note:} The mean accuracy on the whole test set is denoted by Acc. and the one on the test set with only the imbalanced classes by Imb. Acc. } \par}
  	\end{table*}
  	
    \subsubsection{Experiment II}
\begin{figure}
        \begin{center}
            \hspace*{3ex} \scriptsize{Poor} \hspace*{19ex} \scriptsize{Good}
        \end{center}
        \vspace*{-1ex}
        \centering
        \tikz[baseline]{\draw (0,0.55) node [rotate=90] [inner sep=0] {\scriptsize{Locality-}};}
        \tikz[baseline]{\draw (0,0.5) node [rotate=90] [inner sep=0] {\scriptsize{based}};}
        \includegraphics[width=0.13\linewidth]{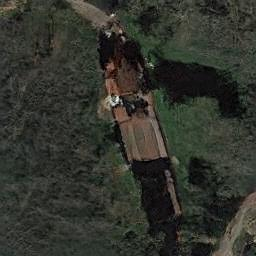}
        \includegraphics[width=0.13\linewidth]{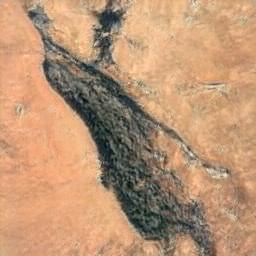}
        \tikz[baseline]{\draw[line width=0.5mm,densely dashed] (1.5ex,0ex)--++(0,1.2) ;}
        \includegraphics[width=0.13\linewidth]{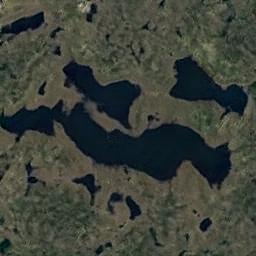}
        \includegraphics[width=0.13\linewidth]{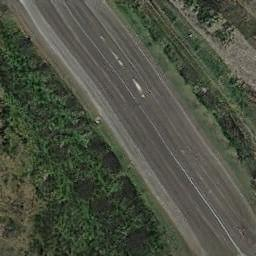}\\[-0.2ex]
        \tikz[baseline]{\draw (0,0.58) node [rotate=90] [inner sep=0] {\scriptsize{PCA-based}};}
        \tikz[baseline]{\draw (0,0.1) node [rotate=90] [inner sep=0.3mm] {\scriptsize{ }};}
        \tikz[baseline]{\draw (0,0.1) node [rotate=90] [inner sep=0] {\scriptsize{ }};}
        \includegraphics[width=0.13\linewidth]{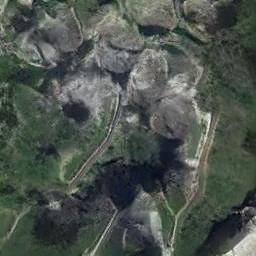}
        \includegraphics[width=0.13\linewidth]{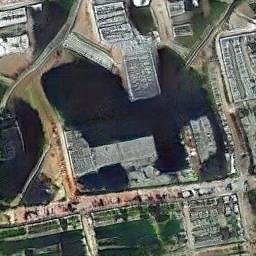}
        \tikz[baseline]{\draw[line width=0.5mm,densely dashed] (1.5ex,0ex)--++(0,1.2) ;}
        \includegraphics[width=0.13\linewidth]{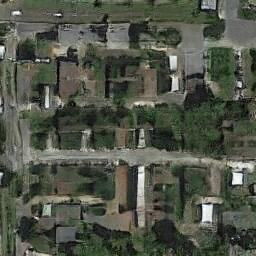}
        \includegraphics[width=0.13\linewidth]{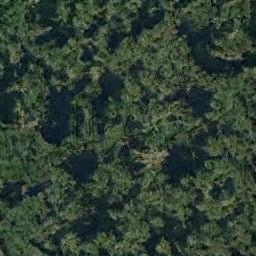}
        \caption{Examples of augmented images that were rejected by ResNet50 in Experiment I. Zoom-in is encouraged.}
        \label{fig:rejected-exp1}
    \end{figure}

    \looseness-1Observing the images that were rejected by ResNet-50 in Experiment I, examples of which can be found in Fig. \ref{fig:rejected-exp1}, it is noticeable that the model rejects many poor results in the case of the PCA-based method. On the contrary, the same model rejects many good results of the LPP-based method. This is mainly due to the fact that the latent codes produced via the proposed method tend to stay within the desired distribution, since LPP preserves the local neighbourhoods of the data. With this experiment we want to test a different probability threshold of 0.5 in the ResNet-50 model for the image scene classification. The training sizes follow the pattern of the previous experiment, which suggests an increase by 5 for Baseline, PCA and LPP-based methods and an increase by 9 for the mixed ones.
    
    Performing this experiment, we expected that the PCA-based method would not benefit from the lower threshold, since it allows more images with artifacts to augment the datasets. This is confirmed by the results of this experiment in comparison to Experiment I: a lower threshold leads to lower accuracy for both methods - with worse results for the PCA-based approach.

    \subsubsection{Experiment III \& IV}
    \looseness-1In the previous two experiments the training size of each dataset variant using the mixed methods is larger that the one created by the other three techniques. The objective in these experiments is to test all the data augmentation techniques using the same settings i.e. increasing the training size by 9, by a) Applying larger magnitudes of $\alpha=\pm 2.5, \pm 3$ on the existing latent codes that augmented the datasets in Experiment I and b) Including new latent codes with the same two magnitudes that we used in Experiment I i.e. $\alpha=\pm 1, \pm 2$.
    
    \looseness-1In both of these cases, the results across all methods are relatively comparable, with the geometric transformations bringing on a slight improvement in almost all dataset variants. This gives rise to the limitations of the context-based methods, which suggest that there is a point up to which we can increase the variability introduced into the imbalanced classes by using only these techniques. However, when we compare the results of Baseline techniques in both experiments presented in Table \ref{table:experiment3+4-baseline}b with the Mixed LPP-based method from Experiment I in Table \ref{table:experiment1-mixedresults}a, we can easily notice that even when the training sizes are the same, the performance of the second method is much better, resulting in higher accuracies. Therefore, using not only geometric transformations to augment a dataset but also context-based ones, can have a great impact on the performance.
    
    \subsubsection{Experiment V}
    \looseness-1To compare the results produced by the PCA and LPP-based methods, we want to examine whether the augmented samples can preserve the class of the initial image or not. In this experiment, we test the magnitudes $\alpha=\pm 0.5, \pm 1$ by only classifying the image produced by SWAGAN and use its label to annotate the augmented results. These images were then used to increase the training size of the datasets by a factor of 5.
    
    Results in Table \ref{table:experiment5-results}c evidently show that our method tends to outperform the PCA-based method in keeping the class of the initial image, while increasing the variability introduced into the imbalanced classes. Both methods' resulting accuracies are still comparable with the ones in Experiment I, considering that in all settings the mean difference of the corresponding accuracies is very small (approximately $3\times10^{-2}$).
    
    \section{Conclusion}
    We presented a locality-preserving  approach for interpreting the latent space of GANs trained on satellite images. Differentiating from recent work that relies on variability-maximization approaches, we show that preserving locality can improve  robustness to artifacts, while also better preserving class information.  Our method leads to improved accuracy in downstream tasks such as remote scene classification, while it can be used in a complementary fashion to traditional geometric augmentation approaches to further improve results.

    \bibliographystyle{IEEEtran}
   	\bibliography{IEEEabrv, bibfile}

\end{document}